\documentclass[letterpaper]{article} 
\usepackage{aaai24}  
\usepackage{times}  
\usepackage{helvet}  
\usepackage{courier}  
\usepackage[hyphens]{url}  
\usepackage{graphicx} 
\usepackage{natbib}  
\usepackage{caption} 
\usepackage{algorithm}
\usepackage{algorithmic}
\usepackage{amsmath,amsfonts}
\usepackage{subcaption}
\usepackage{newfloat}
\usepackage{listings}
\DeclareCaptionStyle{ruled}{labelfont=normalfont,labelsep=colon,strut=off} 
\floatstyle{ruled}
\newfloat{listing}{tb}{lst}{}
\floatname{listing}{Listing}
\title{Automated Hit-frame Detection for Badminton Match Analysis}
\author{
    Yu-Hang Chien,
    Fang Yu
}
\affiliations{

    National Chengchi University, Taipei, Taiwan\\
    moneychien20639@gmail.com, yuf@nccu.edu.tw
}

\begin{document}

\maketitle

\begin{abstract}
Sports professionals constantly under pressure to perform at the highest level can benefit from sports analysis, which allows coaches and players to reduce manual efforts and systematically evaluate their performance using automated tools. This research aims to advance sports analysis in badminton, systematically detecting hit-frames automatically from match videos using modern deep learning techniques. The data included in hit-frames can subsequently be utilized to synthesize players' strokes and on-court movement, as well as for other downstream applications such as analyzing training tasks and competition strategy. The proposed approach in this study comprises several automated procedures like rally-wise video trimming, player and court keypoints detection, shuttlecock flying direction prediction, and hit-frame detection. In the study, we achieved 99\% accuracy on shot angle recognition for video trimming, over 92\% accuracy for applying player keypoints sequences on shuttlecock flying direction prediction, and reported the evaluation results of rally-wise video trimming and hit-frame detection.
\end{abstract}

\section{Introduction}
Badminton, the fastest racquet sport in the world, is played by players of all ages and can be played in singles or doubles configurations \cite{phomsoupha2015science, ramasamy2021kinetic}. Badminton is also a fast-paced and explosive sport that requires athletes' limbs to perform sudden motions such as arm and racquet movements, jumps, lunges, and quick direction changes to deliver various types of strokes, \cite{rambely2005contribution, kuntze2010biomechanical, huang2016kinematic}. By 2022, badminton is expected to have 220 million active players, according to the Badminton World Federation (BWF). As the sport becomes increasingly popular, there are a rising number of research and analysis on it. Stroke and on-court movement sequences are commonly used in badminton sports analysis to show how the match progressed and offer tactical information; hit-frames, or the frames when a player strikes the shuttlecock and alters its flight direction, can be utilized to obtain this information. Additionally, hit-frames possess essential data, including the players' striking location, the shuttlecock's landing zone, and their movements as they return the shuttlecock. This information can be used for further tactical analysis. As a result, such information benefits coaches and players as it aids in improving performance and formulating winning strategies. However, no automated badminton programs can identify hit-frames in raw match videos. By outlining a methodical strategy for automatically identifying hit-frame sequences from publicly available match videos, this work intends to enhance badminton sports analysis. Our method can facilitate downstream applications dependent on information included in hit-frames, such as stroke classification \cite{chu2017badminton}, movement recognition \cite{valldecabres2020players}, and tactical analysis \cite{wang2022modeling}.

\subsection{Related Works}\label{Related Works}
Focusing on trajectory-related studies, \cite{liu2022monotrack} reconstructs the 3D trajectory of shots from unlabeled videos using various neural networks and a per-shot trajectory reconstruction method. \cite{lee2016badminton} demonstrated a model-based trajectory estimation method based on linear regression and SVM. \cite{sun2020tracknetv2} developed TrackNetV2, to forecast the trajectory of the tiny and fast-moving shuttlecock from badminton videos. 

For stroke recognition and prediction research, \cite{wang2016badminton} proposed a badminton stroke recognition system based on body sensors and a two-layer hidden Markov model. \cite{chen2007statistical} described applying 2-D seriate images to obtain statistical data from a badminton match. \cite{chu2017badminton} integrated visual analysis techniques to detect the court and players, classify strokes, and the player's strategy. \cite{ghosh2018towards} proposed a method to analyze badminton broadcast videos by segmenting the points played, tracking and recognizing the players in the points, and commenting on their respective badminton strokes. \cite{careelmont2013badminton} attempted to solve the problem of badminton stroke classification in compressed videos using shuttlecock trajectory. \cite{wang2022shuttlenet} propose a novel Position-aware Fusion of Rally Progress and Player Styles framework that incorporates rally progress and information of the players by two modified encoder-decoder extractors. \cite{wang2022modeling} offered a unified badminton language to describe the shot process and measured the win probability of each shot in badminton matches by considering long-term and short-term dependencies. \cite{wang2022modeling} also introduced a framework with two encoder-decoder extractors and a position-aware fusion network to forecast the possible tactics of players.

As for studies related to footwork and movements, \cite{valldecabres2019design} designed a reliable tool for recording and analyzing the behavior of badminton singles players, then took advantage of the tool and conducted a study that examined the relationship between the elite badminton players' movements and contextual variables (game, round, and match status) \cite{valldecabres2020players}. \cite{abdullahi2017notational} investigated the relationships between singles match strokes and foot movements in African Badminton Championship male badminton players.

These works primarily concentrated on specific components such as trajectory construction, stroke classification, and movement analysis. Moreover, many of them rely on the information within the hit-frames (discussed in \ref{Hit-Frame Applications}), like players' location and the players' motions when they hit the shuttlecock, for further analysis and application. To our knowledge, no previous work predicts the shuttlecock's flying direction with a transformer model, given the keypoints sequence of on-court players. This study provides a systematic way to generate shuttlecock flying direction sequences, then detect hit-frames according to the sequence. Such an approach can facilitate data labeling and downstream applications requiring hit-frames information.

\subsection{Proposed Approaches}
Badminton is a game with a 21-point scoring system. Consequently, it is challenging to collect data from badminton videos all at once. In such a case, we use heuristic methods to divide raw match video into frames based on rallies, forecast the shuttlecock's direction of flight in each frame, and pinpoint hit-frames. Aalgorithm \ref{alg:framework} depicts the process of all the procedures. The hit-frame generating algorithm inputs a badminton frame sequence (video) $V$ and produces a set of hit-frame indices denoted by $Hit$. Lines 1 to 3 define the shot angle constants $O$ (representing $other$) and $H$ (representing $high$). These constants are used to realize the rally-wise video trimming module (described in \ref{Rally-wise Video Trimming}). The variable $previous\_shot\_angle$ is initialized to $O$ to keep track of the shot angle of the previous frame. Two sets are also created: $Kps$ (for storing detected player keypoints) and $Hit$ (for storing detected hit-frame indices), both initially empty. From line 4, the algorithm proceeds to iterate through each frame $frame$ in the input $V$. For each frame, the shot angle is recognized using the function AngleRecognition (described in \ref{Shot Angle Recognition}), and the algorithm follows a series of conditional checks to identify consecutive frames belonging to the same rally. 
\begin{enumerate}
    \item If the angle of the current frame $angle$ matches the angle of the previous frame $previous\_shot\_angle$ and equals to $H$ (like line 6), which indicates the rally is still ongoing, the algorithm detects the keypoints $kps$ of such frames using DetectPlayerKeypoints (elaborated in \ref{Court and Player Keypoints Detection}) and add $kps$ to the $Kps$ set.
    \item If $angle$ matches $previous\_shot\_angle$ and equals $O$ (like line 10), the algorithm will overlook such frame and iterate back to line 4. 
    \item  If $angle$ differs from $previous\_shot\_angle$ and equals $O$ (like line 14), it indicates the end of a rally, the algorithm predicts the flying directions of the shuttlecock using keypoints stored in $Kps$ (elaborated in \ref{Shuttlecock Flying Direction Prediction}) and assigns it to $s$. Then, it proceeds to detect hit-frames $hit$ using HitFrameDetection (illustrated in \ref{Hit-Frame Detection}) and adds it to the $Hit$ set. The $Kps$ set is then cleared, waiting for a new rally to begin.
    \item If $angle$ differs from $previous\_shot\_angle$ and equals $H$ (line 19), it indicates the start of a rally. The algorithm will process such frames like those in lines 8 and 9.
\end{enumerate}
After the algorithm processes all frames in the input sequence, it returns the set $Hit$, which contains detected hit-frames indices of the input video.
\begin{algorithm}[tb]
\caption{Hit-frame generating algorithm.}
\label{alg:framework}
\textbf{Input}: Badminton frame sequence $V$.\\
\textbf{Output}: Hit-frame set $Hit$.
\begin{algorithmic}[1] 
\STATE Let \textbf{const} $O \gets 0, H \gets 1$
\STATE Let $previous\_shot\_angle \gets O$
\STATE Let $Kps \gets \emptyset, Hit \gets \emptyset$
\FORALL{$frame \in V$} 
\STATE $angle \gets \textbf{AngleRecognition}(frame)$
\IF{$angle = previous\_shot\_angle$}
\IF{$shot\_angle = H$}
\STATE $kps \gets \textbf{DetectPlayerKeypoints}(frame)$
\STATE $Kps.\textbf{add}(kps)$
\ELSE
\STATE \textbf{continue}
\ENDIF
\ELSE
\IF{$shot\_angle = O$}
\STATE $s \gets \textbf{ShuttlecockDirPrediction}(Kps)$
\STATE $hit \gets \textbf{HitFrameDetection}(s)$
\STATE $Hit.\textbf{add}(hit)$
\STATE $Kps \gets \emptyset$
\ELSE
\STATE $kps \gets \textbf{DetectPlayerKeypoints}(frame)$
\STATE $Kps.\textbf{add}(kps)$
\ENDIF
\ENDIF
\STATE $previous\_shot\_angle \gets shot\_angle$
\ENDFOR
\STATE \textbf{return} $Hit$
\end{algorithmic}
\end{algorithm}

\subsection{Contribution}
The main contributions of our article are to:
\begin{enumerate}
    \item Utilize the interchange of shot angle to trim rally-wise frames from raw badminton video.
    \item Proposed a novel transformer that predicts shuttlecock direction sequences based on player keypoint sequence.
    \item Designed and developed the first-ever automated hit-frame detection tool to bridge the gap between raw badminton videos and analyzable data.
\end{enumerate}

\section{Shuttlecock Flying Direction Prediction}
This section elaborates on the proposed rally-wise video trimming, human and court keypoint detection, player detection, and shuttlecock flying direction prediction procedures. We adopted broadcast badminton videos from the public BWF YouTube channel as our data source of the dataset we utilized in the following paragraphs.

\subsection{Rally-wise Video Trimming}\label{Rally-wise Video Trimming}
To analyze the video of badminton matches with a 21-point scoring system simultaneously is challenging; we reduced the mission to a point-by-point (rally-by-rally) task and implemented a rally-wise video trimming module to trim all the rally-related frames from the original video. These frame segments can represent player behaviors in each rally. 

\begin{figure}[!t]
\begin{center}
\includegraphics[width=\linewidth]{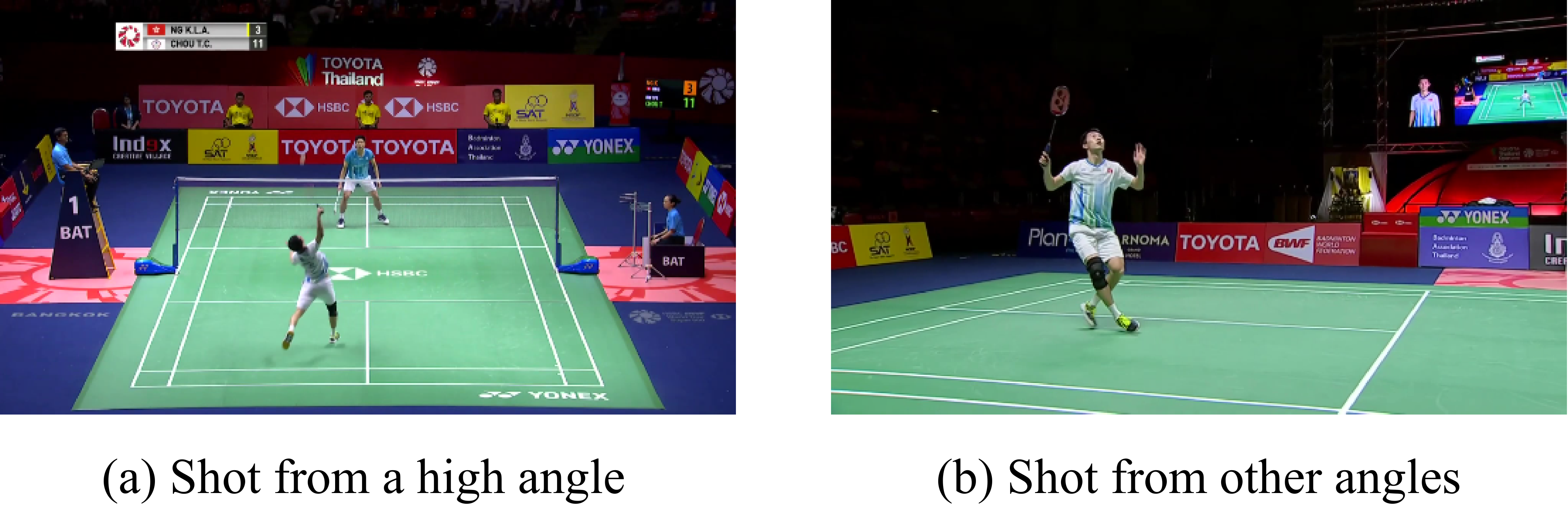}
\end{center}
\caption{Frame shot from different angles.}
\label{Fig:sa}
\end{figure}

We noticed that the match videos were shot from a high angle when the players competed on the court and shot from other angles while replaying highlights or filming other scenes. Figure \ref{Fig:sa} displays the example of frames that were either shot from a high angle or other angles. We formed the issue into a binary classification task and labeled each frame as either \textit{high} (class 1) or \textit{other} (class 0), based on its shot angle. This way, a badminton video can be seen as a long sequence of \textit{high} and \textit{other}. Assuming that the scenes in the rallies are always shot from a high angle, we set the frame as the rally's start frame when the shot angle varies from \textit{other} to \textit{high} and the frame as the rally's end frame when the output varies from \textit{high} to \textit{other}. As a result, the rally-wise video trimming module can generate rally-related frames.

\subsubsection{Shot Angle Recognition}\label{Shot Angle Recognition}
\begin{figure}[!t]
\begin{center}
\includegraphics[width=\linewidth]{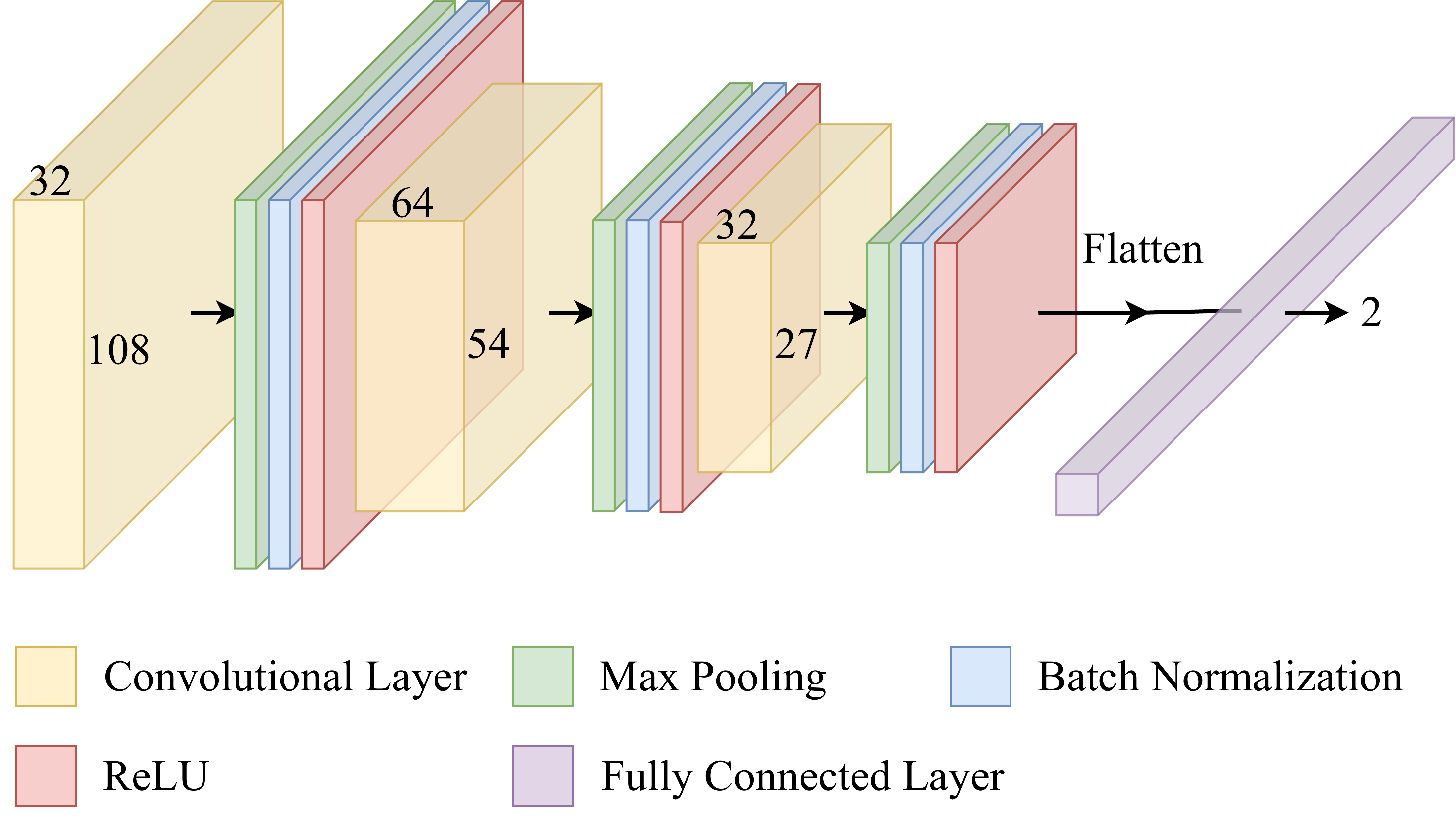}
\end{center}
\caption{The SA-CNN architecture.}
\label{Fig:CNN}
\end{figure}

We adopted the shot angle convolutional neural network (SA-CNN) model, as shown in Figure \ref{Fig:CNN}, to solve the binary classification problem. The dataset we built for the shot angle recognition task comprises 112660 frames collected from 6 unique men's singles matches publicly available on the BWF channel. We divided them into training and testing datasets, with 62970 (4 matches) and 49690 (2 matches) frames, respectively. Additionally, badminton coaches have labeled each frame with one of the two classes,  \textit{high} and \textit{other}. The model's input is expected to be a three-channel RGB image $x$ where $x \in \mathbb{R}^{3 \times 1080 \times 1920}$, and it will be normalized with the ImageNet dataset's means and standard deviations so the pixels are scaled to a range of 0 and 1 \cite{deng2009imagenet}. Other preprocessing steps are defined as follows:
\begin{align*}
    &x^\prime = \text{Resize}(x), x^\prime \in \mathbb{R}^{3 \times 216 \times 384}\\
    &x^{\prime\prime} = \text{CenterCrop}(x^\prime), x^{\prime\prime} \in \mathbb{R}^{3 \times 216 \times 216}\\
    &X = \{x^{\prime\prime}_1 \cdots x^{\prime\prime}_N\}, X \in \mathbb{R}^{N\times 3 \times  216 \times 216}\\
    &\hat{y} = \text{SA-CNN}\left(X\right),  \hat{y} \in \mathbb{R}^{N \times C}
\end{align*}
Where \( \text{Resize}(\cdot) \) is the function to resize the image to the specified dimensions and \( \text{CenterCrop}(\cdot) \) is the function to crop the image at the center to the specified dimensions. At the same time, $N$ and $C$ denote the minibatch size and the number of classes, respectively. Moreover, $X$ is the input batch, and $\hat{y}$ is the model output. The layers in SA-CNN is defined as follows:
\begin{align*}
    &H^{(l)}_{\text{conv}} = \sum_{c=1}^{C^{(l-1)}_{\text{out}}}\sum_{i=1}^{3}\sum_{j=1}^{3} H^{(l-1)}_{\text{out}, c, i, j} W^{(l)}_{c, i, j} + b^{(l)}\\
    &H^{(l)}_{\text{pool}} = \text{MaxPool}\left(H^{(l)}_{\text{conv}}, Pool, Stride\right)\\
    &H^{(l)}_{\text{BN}} = \text{BatchNorm}\left(H^{(l)}_{\text{pool}}\right)\\
    &H^{(l)}_{\text{out}} = \text{ReLU}\left(H^{(l)}_{\text{BN}}\right) = \text{max}\left(0, H^{(l)}_{\text{BN}}\right)
\end{align*}
where \(H^{(l)}_{\text{conv}}\), \(H^{(l)}_{\text{pool}}\), and \(H^{(l)}_{\text{BN}}\) denotes the output of the $l_{th}$ convolutional, maxpooling, and batch normalization layer, respectively. $C^{(l-1)}_{\text{out}}$, $W$, and $b$,  represent the number of input channels, learnable weight matrices, and bias for the $l_{th}$ convolutional layer, respectively. $Pool$ is the pooling window size, while $Stride$ is the pooling stride; both $P$ and $S$ are set as 2. The final output of the convolutional neural network is then flattened and enters a fully connected layer for classification:
\begin{align*}
&H_{\text{flat}} = \text{Flatten}(H^{(L)}_{\text{out}})\\
&H_{\text{FC}} = \text{ReLU}\left(H_{\text{flat}} \cdot W_{\text{FC}} + b_{\text{FC}}\right)\\
&y_{\text{pred}} = \text{argmax}(H_{\text{FC}}), y_{\text{pred}} \in \{high, other\}
\end{align*}
where $L$ stands for the layer count of SA-CNN, \(y_{\text{pred}}\) represents the predicted class, and \(H_{\text{FC}}\) represents the network's output before applying argmax.

The SA-CNN is trained for 20 epochs with an Adam optimizer with the weight decay rate set at constant 0.1 and the learning rate set at 1e-3 but decayed 90\% every six epochs. During the training phase of SA-CNN $N=8, C=2, L=3$ and the loss function $L$ that we used is shown in Eq.\ref{ce}.

\begin{align}\label{ce}
    &L = l( \hat{y}, y) = \{ l_1 \cdots l_n\}^T\\
    &l_n = -\text{log}\frac{\text{exp}\left( \hat{y}_{n, y_n}\right)}{\sum^C_{c=1}\text{exp}\left( \hat{y}_{n, c}\right)}\\
    &l(\hat{y}, y) = \sum^N_{n=1}{l_n}
\end{align}

\subsection{Court and Player Keypoints Detection}\label{Court and Player Keypoints Detection}
After the rally-wise video trimming module generated the frame sequence of a rally, we adopted R-CNN models to detect human and court keypoint sequences. These keypoint sequences showcase the positional and motion information of the players and will be applied in shuttlecock flying direction prediction later.

\subsubsection{Human keypoint Detection}
For human keypoint detection, we adopted the Keypoint R-CNN model provided by PyTorch \cite{paszke2019pytorch}. The Keypoint R-CNN is pre-trained with the COCO dataset \cite{lin2014microsoft}, and its architecture is elaborated in \cite{he2017mask}. Mask R-CNN is a model that extends the Faster R-CNN \cite{girshick2015fast} by adding a branch for predicting segmentation masks on each Region of Interest in parallel with the existing branch for classification and bounding box regression. The Keypoint R-CNN model is based on the Mask R-CNN method and expects its input to be an image of all sizes while the output is a list of dictionaries. Each dictionary provides the predicted box, label, confidence score, and an instance's keypoints. In our case, we only utilized the instance's keypoints, which include the following 17 human parts: \textit{nose, left eye, right eye, left ear, right ear, left shoulder, right shoulder, left elbow, right elbow, left wrist, right wrist, left hip, right hip, left knee, right knee, left ankle, right ankle.} The detection process can be defined as:
\begin{equation*}
    I_{rally} \in \mathbb{R}^{1080 \times 1920}
\end{equation*}
\begin{equation*}\label{}
   K = \text{KeypointRCNN}(I_{rally}), K \in \mathbb{R}^{M \times 17 \times 2}
\end{equation*}
where $I_{rally}$ denotes the rally-related frame, $K$ denotes the detected human keypoint coordinates, and $M$ represents the count of humans detected.

\subsubsection{Court Keypoints Detection}
\begin{figure*}[!t]
\centering
\begin{subfigure}{2.2in}
    \includegraphics[width=2.2in]{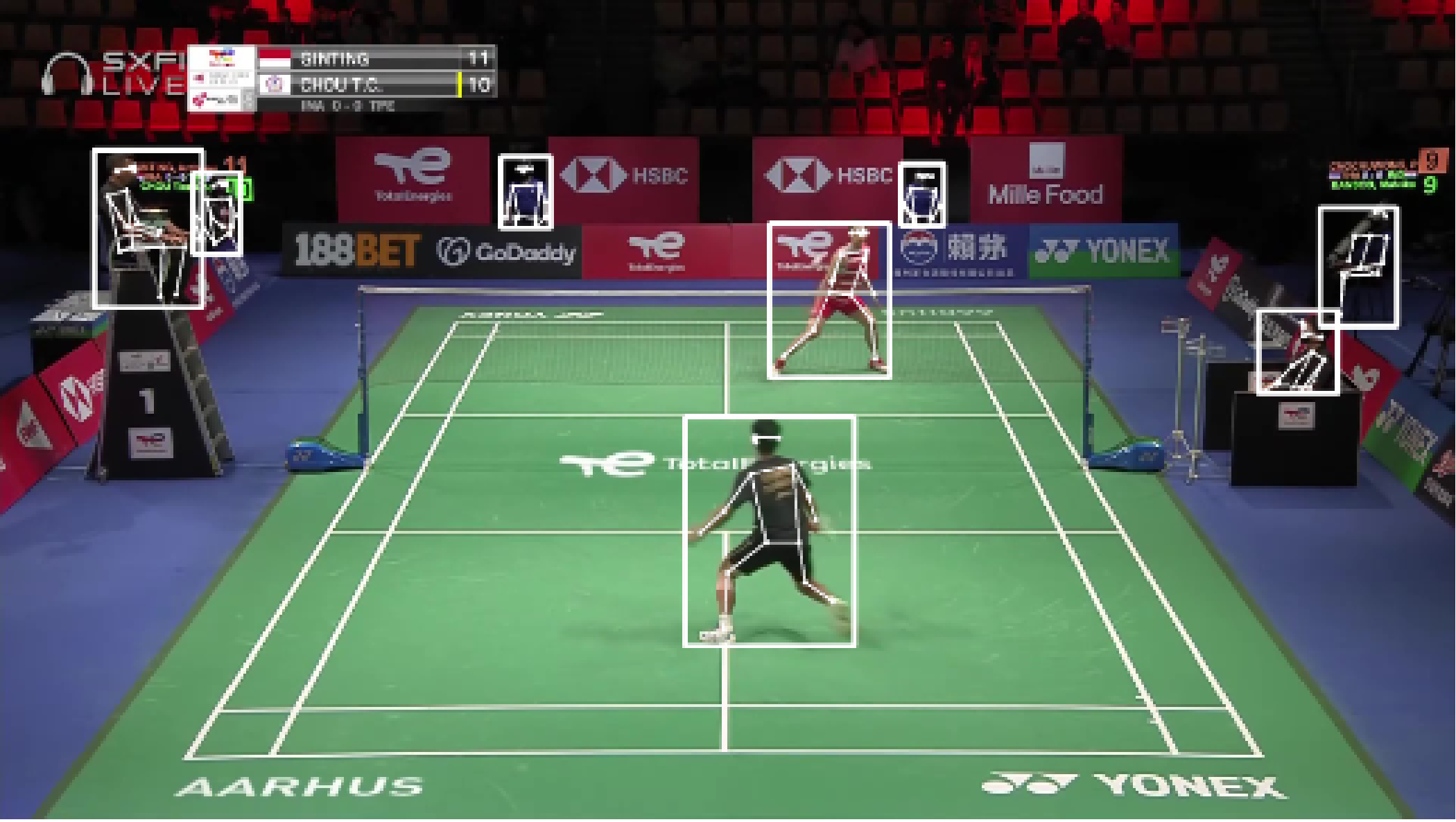}
    \label{unfiltered}\caption{}
\end{subfigure}
\hfil
\begin{subfigure}{2.2in}
    \includegraphics[width=2.2in]{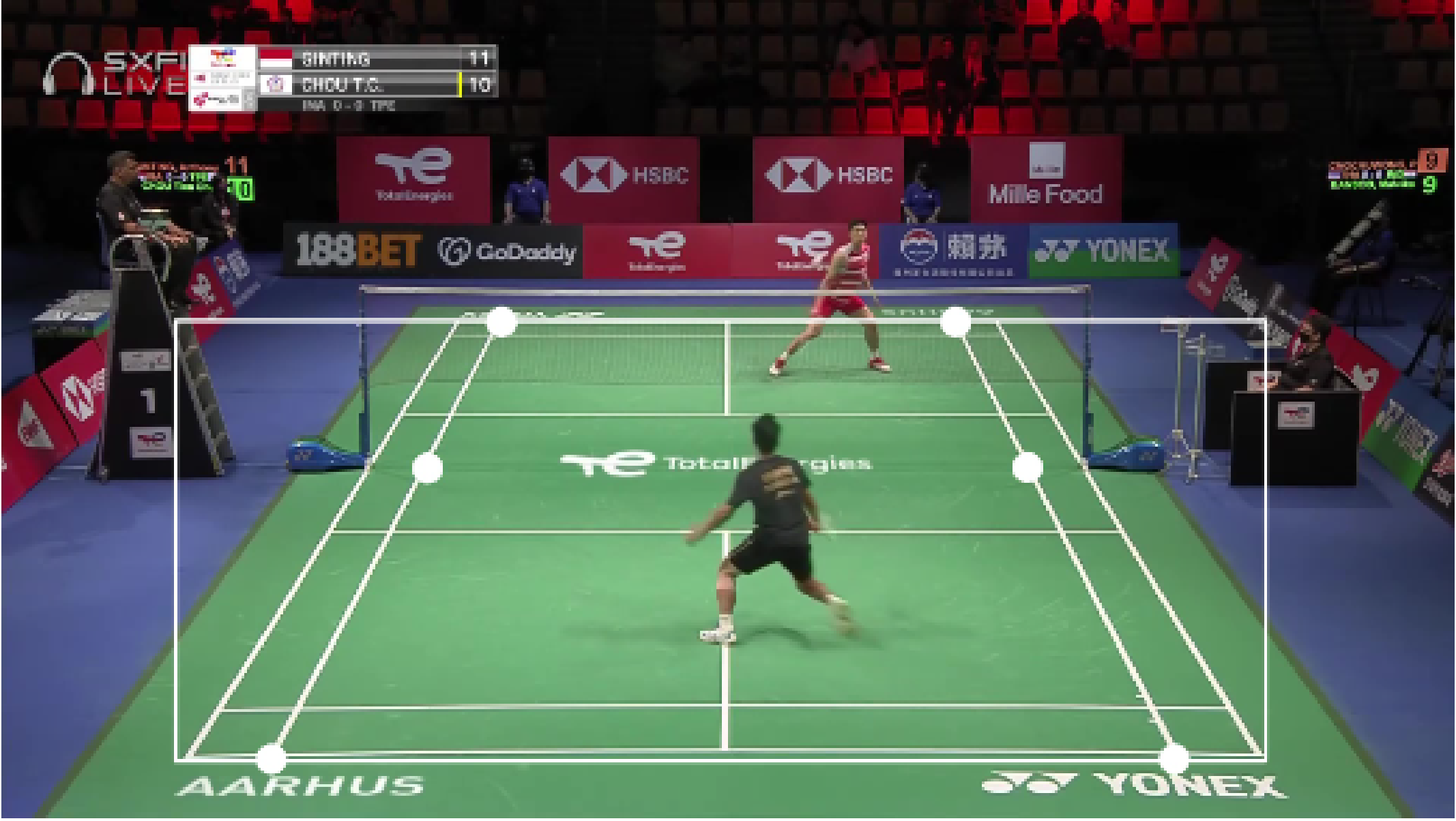}
    \label{court keypoints}\caption{}
\end{subfigure}
\hfil
\begin{subfigure}{2.2in}
    \includegraphics[width=2.2in]{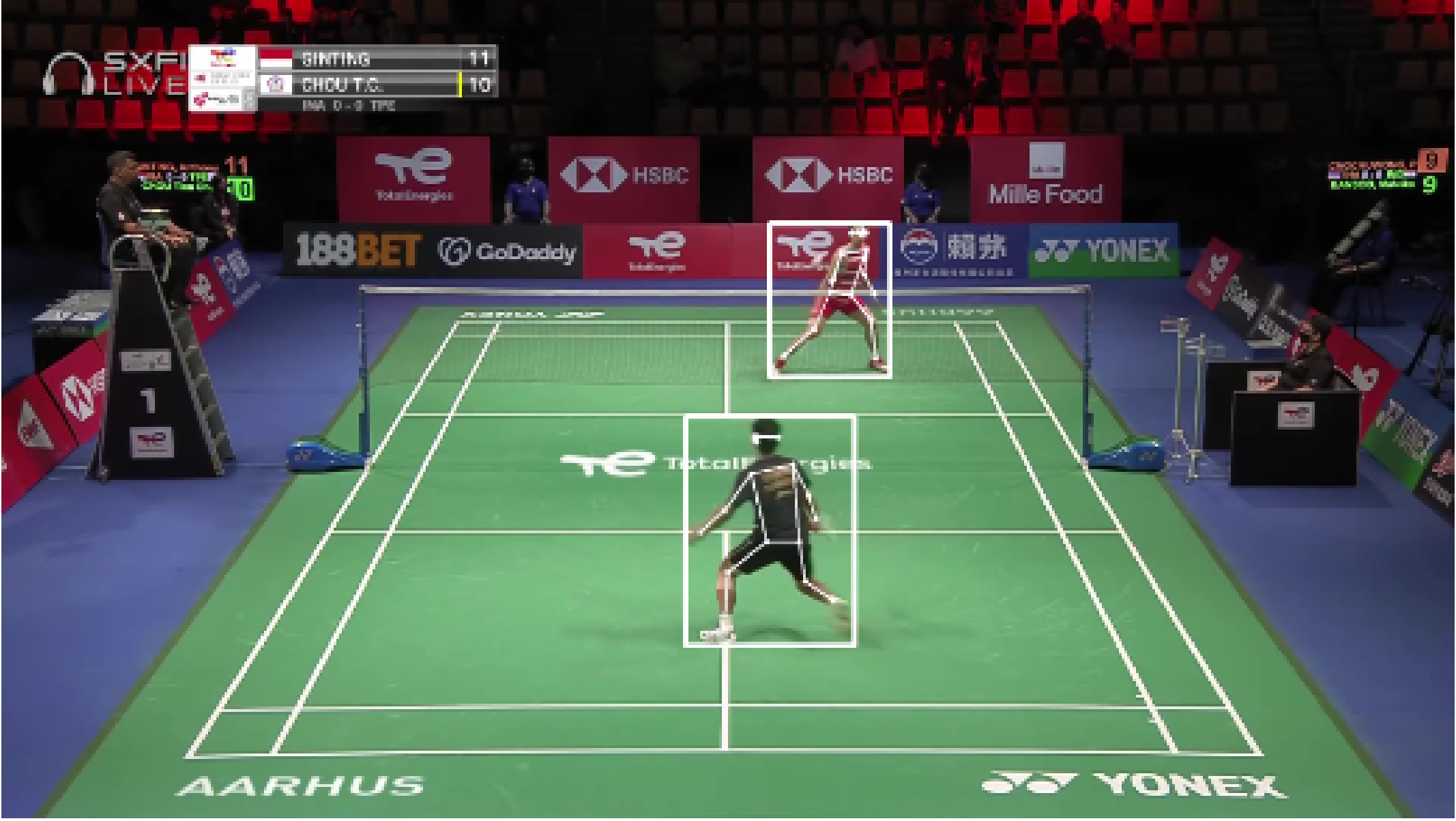}
    \label{filtered}\caption{}
\end{subfigure}
\caption{(a) Illustrates the Keypoint R-CNN model’s output without filtering the instances. (b) shows the output of the Court R-CNN model. (c) Illustrates the outcome of the integrated method of Keypoint R-CNN and Court R-CNN.}
\label{Fig:filtered_output}
\end{figure*}

As Keypoint R-CNN inevitably returns all the humans it detected (as shown in Figure \ref{Fig:filtered_output}(a)), we trained a Court R-CNN for court keypoints detection to filter players’ information. The output format of the Court R-CNN is similar to that of the Keypoint R-CNN, except it predicted only six keypoints, including \textit{upper right, upper left, middle right, middle left, bottom right,} and \textit{bottom left} points (as shown in Figure \ref{Fig:filtered_output}(b)). Since we focus only on men’s singles matches, the detection target of the model will be the singles court. 
\begin{equation*}
   K_{court} = \text{CourtRCNN}(I_{rally}), K_{court} \in \mathbb{R}^{1 \times 6 \times 2}
\end{equation*}
where $K_{court}$ stands for the court's keypoints coordinates.
\subsubsection{Player Keypoints Detection}
We detect player keypoints by integrating the two R-CNN models; the output of the Court R-CNN model is exploited to examine the ankle coordinates of the Keypoint R-CNN output instances. To determine whether the detected instance is a player, we examined its ankle coordinates and ensured it locates within the region defined by the court keypoints. Figure \ref{Fig:filtered_output}(c) displays the filtered output.
\begin{equation*}
   K_{player} = \text{Filter}(K, K_{court}), K_{player} \in \mathbb{R}^{2 \times 17 \times 2},
\end{equation*}
where $K_{player}$ stands for the players' keypoints and \( \text{Filter}(\cdot) \) is the function that takes the court and human's keypoints as its input and outputs the on-court player's keypoints.

\subsection{Shuttlecock Flying Direction Prediction}\label{Shuttlecock Flying Direction Prediction}
Identifying the shuttlecock's flying direction is essential for downstream analysis since it implies information about when the players hit the shuttlecock. In a badminton rally, shots will occur several times until it ends, when a fault happens, or when the shuttlecock hits the floor. Therefore, we conclude the flying status of the shuttlecock into the following three state classes: \textit{steady}, \textit{flying toward the bottom court}, and \textit{flying toward the upper court}, then formulate the shuttlecock flying direction prediction as a sequence labeling task with tokens $S$, $B$, and $U$ representing the three classes respectively. We assume the player keypoint sequence is essential in predicting the shuttlecock's flying direction. Consequently, we proposed a novel transformer model that leverages keypoint sequences to predict shuttlecock flying direction sequences. 

For the shuttlecock flying direction prediction task, we built a keypoint sequence dataset (KSeq dataset) with 12 unique men's singles matches from the BWF channel as the data source. Ten matches were used to train the transformer model, while two were for testing. The sequence data in the KSeq dataset comprises multiple frame-wise keypoint pairs (2 players). Table \ref{KSeq} shows the information of KSeq dataset. In total, the dataset is composed of 1134 rally-wise keypoint sequences. For the ground truth, we invited several badminton experts to label each keypoint pair with one of the mentioned three classes according to the shuttlecock flying direction in the corresponding video frame. Furthermore, padding token (token $Pad$) is applied to ensure each keypoint sequence has the same length during training, and the keypoints will be normalized with the dataset's average and standard deviation before entering the transformer.

\begin{table}
\caption{Description of KSeq dataset.}
\label{KSeq}
\centering
\begin{tabular}{lccc}
\hline
            &  Matches  &  Keypoint Sequences  &  Keypoint Pairs\\
\hline
      Train &  10       &  881       &  120147    \\
      Test  &  2        &  253        &  37288    \\
      Total &  12       &  1134       &  157435   \\
\hline
\end{tabular}
\end{table}

\begin{figure}[!t]
\begin{center}
\includegraphics[width=\linewidth]{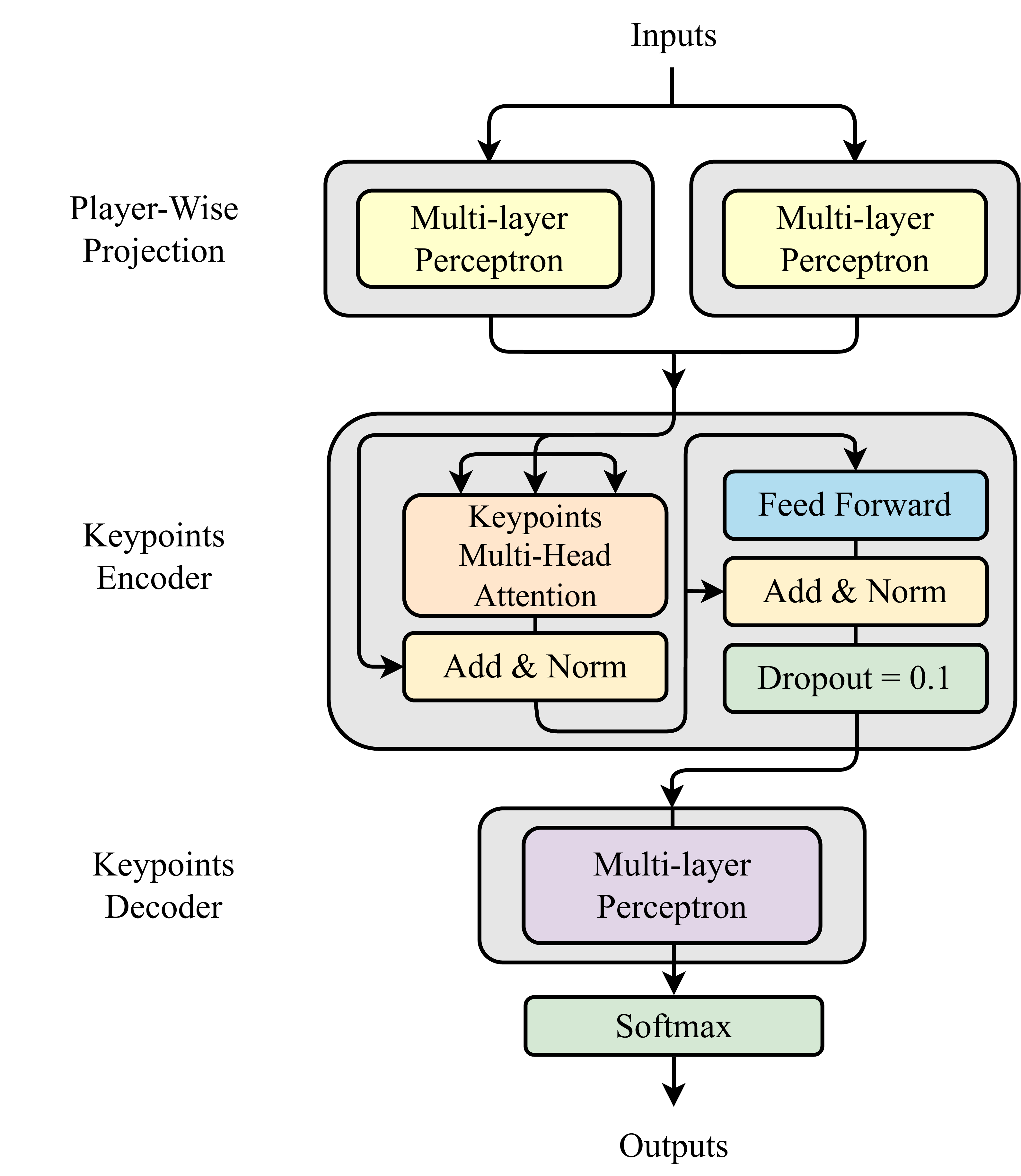}
\end{center}
\caption{ The architecture of the adopted transformer model.}
\label{Fig:transformer}
\end{figure}

We trained the transformer for 100 epochs with an Adam optimizer, whose learning rate is set as 1e-5 and decayed 90\% after the seventieth epoch. The training process is defined as follows:
\begin{align*}
    &x^t = (K_{player(1)} \cdots K_{player(F)})\\
    &X^t = (x^t_1 \cdots x^t_N), x^t \in \mathbb{R}^{F \times 2 \times 17 \times 2}\\
    &\hat{y^t} = \text{Transformer}(X^t), \hat{y^t} \in \mathbb{R}^{N\times F \times C}
\end{align*}
where $x^t$ stands for the rally-wise keypoint sequences, $X^t$ represents the training data batch, $y^t$ is the shuttlecock direction sequence, and $F$ represents the frame sequence length. The loss function $L^t$ for transformer training is defined as follows: 
\begin{align}
    &y^t=\{y^t_{1} \cdots y^t_{N}\}, y^t_{n} \in\{S, B, U, Pad\}\\
    &L^t = l^t(\hat{y^t}, y^t) = \{ l^t_1 \cdots l^t_n\}^T\\
    &Ignore(x) = 
    \begin{cases}
    1, & \text{if } x \neq \text{ignore index} \\
    0, & \text{else}
    \end{cases}\\
    &l^t_n = -\text{log}\left(\frac{\text{exp}\left(\hat{y^t}_{n, y^t_n}\right)}{\sum^C_{c=1}\text{exp}\left(\hat{y^t}_{n, c}\right)}\right)\cdot Ignore(y^t_n)\\
    &l^t(\hat{y^t}, y^t) = \sum^N_{n=1}\frac{l^t_n}{\sum^N_{n=1}Ignore(y^t_n)}
\end{align}
In the loss function $L^t$, $N$ spans the batch dimension as well as $F$. During training, $N = 1, F = 600, C = 4$, and the padding token $Pad$ is set as the ignore index, so it is ignored and does not contribute to the input gradient when calculating the loss. Figure \ref{Fig:transformer} depicts the transformer's structure.

\subsubsection{Player-Wise Projection}
A player-wise projection layer is built to initially project the input keypoints to a higher dimension before feeding the transformer sequences of numerous pairs of keypoints. The player-wise projection layer comprises two multi-layer perceptron blocks with two fully connected layers. Before entering the layer, the keypoint data of the two players will be momentarily separated, and following the projection procedure, they will be concatenated. In our experiment, the transformer model with player-wise projection layers outperforms the model without it, supporting our hypothesis that this mechanism helps separate the information of the two players rather than considering them as a single identity. The player-wise projection layer is defined as follows:
\begin{align*}
    &FC(x) = \text{ReLU}\left(W_{FC} \cdot x + b_{FC}\right)\\
    &H^{(i)}_{\text{player}} = \text{{FC}}\left(\text{{FC}}\left(x^t[:, i, :, :]\right)\right), i \in \{1, 2\}\\
    &x^t_{\text{projected}} = \text{{Concatenate}}\left(H^{(1)}_{\text{player}}, H^{(2)}_{\text{player}}\right)
\end{align*}
where $W_{FC}$, $b_{FC}$ is the learnable weight matrices and bias of the fully connected layer, $H^{(i)}_{\text{player}}$ denotes the output of the $i_{th}$ player-wise projection layer before concatenation, and $x^t_{\text{projected}}$ represents the projected keypoint sequence tensor.

\subsubsection{keypoint Encoder and Decoder}
The encoder stack comprises several identical layers, each built based on the architecture described in \cite{vaswani2017attention} and consists of two sub-layers. The first is a multi-head self-attention layer; the second is a position-wise fully connected feed-forward network. Around the two sub-layers, residual connection is proposed, followed by layer normalization and dropout. The model's outputs from all sub-layers have identical dimensions to facilitate these residual connections. The transformer model we built features eight encoder layers and eight heads for the attention mechanism; the dimension of the feed-forward layer is 2048.

\subsection{Hit-Frame Detection}\label{Hit-Frame Detection}
Hit-frame, which are defined as the frame that a player hits the shuttlecock and thus changes its flying direction, can be utilized to retrieve data for downstream analysis since they possess information about the players when they hit the shuttlecock. Such hit-frames can be identified by detecting token shifts in the shuttlecock flying direction sequence. Algorithm \ref{alg:hit-frame} depicts how we devised hit-frame detection. 

Lines 1 to 3, the algorithm starts by defining three constants: $S$, $B$, and $U$, which represent the states of the shuttlecock flying direction as follows: \textit{steady}, \textit{flying toward the bottom court}, and \textit{flying toward the upper court}. The variable $previous\_direction$ is initialized to $S$ to keep track of the previous flying direction encountered during the iteration. The set $hit$ is also created and initialized as empty, where the identified hit-frame indices will be stored. The algorithm then enters a loop that iterates over each shuttlecock flying direction in the input sequence $s$. If the current direction differs from the previous direction within the loop, the shuttlecock's flying direction has changed. The algorithm then checks for specific cases of transitions to detect hit-frames.
\begin{enumerate}
    \item If the previous direction was $S$ and the current direction is $B$ or $U$, it tells that a player starts the rally; therefore, the algorithm identifies this frame as a hit-frame. It calls FrameIndex to get the frame index $h$ corresponding to the current direction and adds it to the set $hit$.
    \item If the previous direction was $B$ and the current direction is $U$, or vice versa, it indicates the player at the bottom court strikes the shuttlecock, causing a flying direction transition (or vice versa). The algorithm calls FrameIndex to obtain the frame index $h$ corresponding to the current direction and adds it to the set $hit$.
\end{enumerate}
The algorithm then updates the $previous\_direction$ variable to the current direction before moving on to the next iteration. After all directions in the sequence have been processed, the algorithm returns the set $hit$ containing the detected hit-frame indices, representing potential badminton shots where the shuttlecock changes its flying direction.

\begin{algorithm}[tb]
\caption{Hit-Frame detecting algorithm.}
\label{alg:hit-frame}
\textbf{Input}:  Shuttlecock flying direction sequence $s$.\\
\textbf{Output}: Hit-frames set $hit$.
\begin{algorithmic}[1] 
\STATE Let \textbf{const} $S \gets 0, B \gets 1, U \gets 2$
\STATE Let $previous\_direction \gets S$
\STATE Let $hit \gets \emptyset$
\FORALL{$direction \in s$} 
\IF{$previous\_direction \neq direction$}
\IF{$previous\_direction = S$}
\STATE $h \gets \textbf{FrameIndex}(direction)$
\STATE $hit.\textbf{add}(h)$
\ELSIF{$previous\_direction = B$}
\IF{$direction = U$}
\STATE $h \gets \textbf{FrameIndex}(direction)$
\STATE $hit.\textbf{add}(h)$
\ENDIF
\ELSIF{$previous\_direction = U$}
\IF{$direction = B$}
\STATE $h \gets \textbf{FrameIndex}(direction)$
\STATE $hit.\textbf{add}(h)$
\ENDIF
\ENDIF
\ENDIF
\STATE $previous\_direction \gets direction$
\ENDFOR
\STATE \textbf{return} $hit$.
\end{algorithmic}
\end{algorithm}

\section{Experiments}
Our research intends to enhance sports analysis in badminton by systematically detecting hit-frames from match videos. We seek to address three key research questions: 
\begin{enumerate}
    \item Can rallies be precisely extracted from match videos?
    \item Can shuttlecock flying direction sequence be transformed from the sequence of keypoints of players in frames? 
    \item Can hit-frames be precisely identified through flying direction changes?
\end{enumerate}
We will answer these questions in the following paragraphs, and the assessment and experiment of the SA-CNN, rally-wise video trimming module, transformer, and hit-frame detecting module are the main topics of this section.

\subsection{SA-CNN}
We adopted the testing dataset mentioned in \ref{Shot Angle Recognition} to evaluate SA-CNN. Such dataset comprising of two matches (49690 frames), and all of its frames are labeled with one of the two classes, \textit{high} and \textit{other}. The confusion matrix for the evaluation is displayed in Table \ref{sa conf}. We can observe that SA-CNN has remarkable performance and great accuracy for identifying shot angles.
\begin{table}
\caption{Confusion matrix of the SA-CNN.}
\label{sa conf}
\centering
\begin{tabular}{cccc}
\hline
    Accuracy  &  Precision  &  Recall  &  F1-Score \\
\hline
    0.9976    &  0.9903     &  1.0     &  0.9951   \\
\hline
\end{tabular}
\end{table}

\subsection{Rally-Wise Video Trimming}
While the frames were inputted sequentially into SA-CNN, a continuous sequence of \textit{high} and \textit{other} will be outputted. In this study, we evaluated our rally-wise video trimming module with $\text{S}^2$-Labeling, a shot-by-shot labeling dataset that consists of six badminton matches \cite{huang2022s}; however, we could only find four available video sources of the dataset so those accessible videos will be used for evaluation. The $\text{S}^2$-Labeling dataset provides the rally's start and end frame number, which we adopted in the evaluation stage. Table \ref{vt st} shows the number of rallies trimmed and the actual rally count, while Table \ref{vt conf} shows the confusion matrix of the SA-CNN.
\begin{table}
\caption{Numbers of correctly trimmed, extra trimmed, miss trimmed videos and the actual rally counts.}
\label{vt st}
\centering
\begin{tabular}{ccccc}
\hline
    Correct  &  Extra  &  Missed  &  Total Trimmed & Actual  \\
\hline
    287      &  33     &  36      &  320           & 323     \\
\hline
\end{tabular}
\end{table}
\begin{table}
\caption{Confusion matrix of the rally-wise video trimming module.}
\label{vt conf}
\centering
\begin{tabular}{cccc}
\hline
    Accuracy  &  Precision  &  Recall  &  F1-Score \\
\hline
    0.8062    &  0.8969     &  0.8885  &  0.8927   \\
\hline
\end{tabular}
\end{table}
From Tables \ref{vt st} and \ref{vt conf}, we acknowledged that the video trimming module could not trim all rally-related videos. Such a limitation is due to the varying shooting forms of broadcast badminton videos. This restriction is inevitable as the broadcasting department aims to enhance the audience's viewing experience. However, most of the time, the rally-wise video trimming module can still trim rally-related frames from raw match videos and achieve high precision and recall value.

\subsection{Transformer}
With the transformer model, player keypoint sequences can be transformed into shuttlecock flying direction sequences comprising \textit{steady} (token $S$), \textit{flying toward the bottom court} (token $B$), and \textit{flying toward the upper court} (token $U$). Therefore, for transformer evaluation, we will calculate the confusion matrix of the predicted tokens with the KSeq testing dataset. The evaluation result is shown in Table \ref{tran conf}.
\begin{table}
\caption{Confusion matrix of the transformer model.}
\label{tran conf}
\centering
\begin{tabular}{ccccc}
\hline
    Token  &  Accuracy  &  Precision  &  Recall  &  F1-Score \\
\hline
    $S$      &  0.9717    &  0.9798     &  0.9410  &  0.9600   \\
    $B$      &  0.9275    &  0.8879     &  0.8976  &  0.8927   \\
    $U$      &  0.9295    &  0.8824     &  0.9090  &  0.8955   \\
\hline
\end{tabular}
\end{table}
Table \ref{tran conf} shows that the transformer model performs better when predicting token $S$. Overall, the transformer can robustly tackle the shuttlecock's flying direction prediction task.

\subsection{Hit-Frame Detection}
Since we have trained the transformer model with only men's singles matches, for the evaluation of the hit-frame detection module, we exploited two men's singles matches from the $\text{S}^2$-Labeling dataset (the other two matches are women's singles matches) and the KSeq testing dataset in the evaluation stage; plus, the frames per second (FPS) of match videos are all 30. We assess the hit-frame detection module within three conditions: $\pm5, \pm15, \pm25$, represent the range of the difference between the predicted hit-frame's number and the actual hit-frame's number. If the difference is within the range, the prediction will be considered correct. For instance, condition $\pm 15$ is defined as:
$$\forall n \in \mathbb{Z}, \text{ if } n \in n_{\text{hit}}, P \text{ is correct if } P \in (n - 15, n + 15)$$
where $n$ represents the frame number, $n_{\text{hit}}$ denotes the set of actual hit-frame numbers, and $P$ is the prediction. The result of hit-frame detection is shown in Table \ref{hf conf yc}. From Table \ref{hf conf yc}, we can infer that the hit-frame detecting module can distinguish most hit-frames within the range $\pm 15$.

\begin{table}
\caption{Confusion matrix of the hit-frame detecting module.}
\label{hf conf yc}
\centering
\begin{tabular}{ccccc}
\hline
    Condition &  Accuracy  &  Precision  &  Recall  &  F1-Score \\
\hline
    $\pm5$    &  0.9585    &  0.5454     &  0.7721  &  0.6392   \\
    $\pm15$   &  0.9765    &  0.6787     &  0.9608  &  0.7955   \\
    $\pm25$   &  0.9782    &  0.6915     &  0.9790  &  0.8105   \\
\hline
\end{tabular}
\end{table}

\section{Hit-Frame Applications} \label{Hit-Frame Applications}
Numerous current studies on badminton involve hit-frames. For example, the badminton observational tool validated by \cite{valldecabres2020players} defined on-court movement as a movement done by the observed player, considering the starting and ending zones. The starting zone refers to the quadrant where the observed player is when the opponent hits the shuttlecock, and the ending zone refers to the quadrant where the observed player hits the shuttle back to the opponent's court. Moreover, \cite{chu2017badminton} classify the stroke a player invokes based on his/her pose when he/she hits the shuttlecock. \cite{wang2021exploring} introduce a badminton language BSLR to describe the process of the shot. The BSLR data format includes an essential attribute: timestamp, which is the shot's hitting time (in seconds). Our hit-frame detection framework can obtain such an attribute; therefore, downstream applications based on BSLR like \cite{wang2022modeling} and \cite{wang2022stroke}, which captures a shot-by-shot sequence in a badminton rally, can be realized through our framework. 

\section{Conclusion}
Badminton analysis has predominantly relied on manual methods or coach records. Therefore, scalable, automated information extraction is crucial for the training of AI models. In this paper, we developed an automated information extraction model that captures rally-by-rally information from publicly available badminton videos, bridging the gap between videos and analyzable information. We trimmed the rally frame sequences from raw badminton match videos based on the interchange of shot angle, detected hit-frames, which possess abundant player information, with the shifts in shuttlecock flying direction, and achieved 81\%, 96\% accuracy, respectively. Our approach can further support AI-based applications like stroke classification, movement recognition, and tactical analysis. The badminton community can more effectively assess badminton matches from hundreds of publicly available BWF films with the aid of the hit-frame detection framework. Our code and datasets will be made available to the general public. 

\bibliography{aaai24}

\begin{thebibliography}{27}
\providecommand{\natexlab}[1]{#1}

\bibitem[{Abdullahi and Coetzee(2017)}]{abdullahi2017notational}
Abdullahi, Y.; and Coetzee, B. 2017.
\newblock Notational singles match analysis of male badminton players who
  participated in the African Badminton Championships.
\newblock \emph{International Journal of Performance Analysis in Sport},
  17(1-2): 1--16.

\bibitem[{Careelmont(2013)}]{careelmont2013badminton}
Careelmont, S. 2013.
\newblock Badminton shot classification in compressed video with baseline
  angled camera.
\newblock \emph{Master esis, University of Ghent}.

\bibitem[{Chen and Wang(2007)}]{chen2007statistical}
Chen, B.; and Wang, Z. 2007.
\newblock A statistical method for analysis of technical data of a badminton
  match based on 2-D seriate images.
\newblock \emph{Tsinghua science and technology}, 12(5): 594--601.

\bibitem[{Chu and Situmeang(2017)}]{chu2017badminton}
Chu, W.-T.; and Situmeang, S. 2017.
\newblock Badminton video analysis based on spatiotemporal and stroke features.
\newblock In \emph{Proceedings of the 2017 ACM on international conference on
  multimedia retrieval}, 448--451.

\bibitem[{Deng et~al.(2009)Deng, Dong, Socher, Li, Li, and
  Fei-Fei}]{deng2009imagenet}
Deng, J.; Dong, W.; Socher, R.; Li, L.-J.; Li, K.; and Fei-Fei, L. 2009.
\newblock Imagenet: A large-scale hierarchical image database.
\newblock In \emph{2009 IEEE conference on computer vision and pattern
  recognition}, 248--255. Ieee.

\bibitem[{Ghosh, Singh, and Jawahar(2018)}]{ghosh2018towards}
Ghosh, A.; Singh, S.; and Jawahar, C. 2018.
\newblock Towards structured analysis of broadcast badminton videos.
\newblock In \emph{2018 IEEE Winter Conference on Applications of Computer
  Vision (WACV)}, 296--304. IEEE.

\bibitem[{Girshick(2015)}]{girshick2015fast}
Girshick, R. 2015.
\newblock Fast r-cnn.
\newblock In \emph{Proceedings of the IEEE international conference on computer
  vision}, 1440--1448.

\bibitem[{He et~al.(2017)He, Gkioxari, Doll{\'a}r, and Girshick}]{he2017mask}
He, K.; Gkioxari, G.; Doll{\'a}r, P.; and Girshick, R. 2017.
\newblock Mask r-cnn.
\newblock In \emph{Proceedings of the IEEE international conference on computer
  vision}, 2961--2969.

\bibitem[{Huang et~al.(2016)Huang, Huang, Chung, and Tsai}]{huang2016kinematic}
Huang, K.-S.; Huang, C.; Chung, S.~S.; and Tsai, C.-L. 2016.
\newblock Kinematic analysis of three different badminton backhand overhead
  strokes.
\newblock In \emph{ISBS-Conference Proceedings Archive}.

\bibitem[{Huang et~al.(2022)Huang, Huang, Lee, {\.I}k, and Wang}]{huang2022s}
Huang, Y.-H.; Huang, Y.-C.; Lee, H.~S.; {\.I}k, T.-U.; and Wang, C.-C. 2022.
\newblock S 2-Labeling: Shot-By-Shot Microscopic Badminton Singles Tactical
  Dataset.
\newblock In \emph{2022 23rd Asia-Pacific Network Operations and Management
  Symposium (APNOMS)}, 1--6. IEEE.

\bibitem[{Kuntze, Mansfield, and Sellers(2010)}]{kuntze2010biomechanical}
Kuntze, G.; Mansfield, N.; and Sellers, W. 2010.
\newblock A biomechanical analysis of common lunge tasks in badminton.
\newblock \emph{Journal of sports sciences}, 28(2): 183--191.

\bibitem[{Lee(2016)}]{lee2016badminton}
Lee, C.-L. 2016.
\newblock Badminton Shuttlecock Tracking and 3D Trajectory Estimation From
  Video.

\bibitem[{Lin et~al.(2014)Lin, Maire, Belongie, Hays, Perona, Ramanan,
  Doll{\'a}r, and Zitnick}]{lin2014microsoft}
Lin, T.-Y.; Maire, M.; Belongie, S.; Hays, J.; Perona, P.; Ramanan, D.;
  Doll{\'a}r, P.; and Zitnick, C.~L. 2014.
\newblock Microsoft coco: Common objects in context.
\newblock In \emph{Computer Vision--ECCV 2014: 13th European Conference,
  Zurich, Switzerland, September 6-12, 2014, Proceedings, Part V 13}, 740--755.
  Springer.

\bibitem[{Liu and Wang(2022)}]{liu2022monotrack}
Liu, P.; and Wang, J.-H. 2022.
\newblock MonoTrack: Shuttle trajectory reconstruction from monocular badminton
  video.
\newblock In \emph{Proceedings of the IEEE/CVF Conference on Computer Vision
  and Pattern Recognition}, 3513--3522.

\bibitem[{Paszke et~al.(2019)Paszke, Gross, Massa, Lerer, Bradbury, Chanan,
  Killeen, Lin, Gimelshein, Antiga et~al.}]{paszke2019pytorch}
Paszke, A.; Gross, S.; Massa, F.; Lerer, A.; Bradbury, J.; Chanan, G.; Killeen,
  T.; Lin, Z.; Gimelshein, N.; Antiga, L.; et~al. 2019.
\newblock Pytorch: An imperative style, high-performance deep learning library.
\newblock \emph{Advances in neural information processing systems}, 32.

\bibitem[{Phomsoupha and Laffaye(2015)}]{phomsoupha2015science}
Phomsoupha, M.; and Laffaye, G. 2015.
\newblock The science of badminton: game characteristics, anthropometry,
  physiology, visual fitness and biomechanics.
\newblock \emph{Sports medicine}, 45: 473--495.

\bibitem[{Ramasamy et~al.(2021)Ramasamy, Usman, Sundar, Towler, and
  King}]{ramasamy2021kinetic}
Ramasamy, Y.; Usman, J.; Sundar, V.; Towler, H.; and King, M. 2021.
\newblock Kinetic and kinematic determinants of shuttlecock speed in the
  forehand jump smash performed by elite male Malaysian badminton players.
\newblock \emph{Sports biomechanics}, 1--16.

\bibitem[{Rambely and Osman(2005)}]{rambely2005contribution}
Rambely, A.~S.; and Osman, N. A.~A. 2005.
\newblock The contribution of upper limb joints in the development of racket
  velocity in the badminton smash.
\newblock In \emph{ISBS-Conference Proceedings Archive}.

\bibitem[{Sun et~al.(2020)Sun, Lin, Chuang, Hsu, Yu, Chung, and
  {\.I}k}]{sun2020tracknetv2}
Sun, N.-E.; Lin, Y.-C.; Chuang, S.-P.; Hsu, T.-H.; Yu, D.-R.; Chung, H.-Y.; and
  {\.I}k, T.-U. 2020.
\newblock TrackNetV2: Efficient shuttlecock tracking network.
\newblock In \emph{2020 International Conference on Pervasive Artificial
  Intelligence (ICPAI)}, 86--91. IEEE.

\bibitem[{Valldecabres et~al.(2020)Valldecabres, Casal, Chiminazzo, and
  De~Benito}]{valldecabres2020players}
Valldecabres, R.; Casal, C.~A.; Chiminazzo, J. G.~C.; and De~Benito, A.~M.
  2020.
\newblock Players’ on-court movements and contextual variables in badminton
  world championship.
\newblock \emph{Frontiers in Psychology}, 11: 1567.

\bibitem[{Valldecabres et~al.(2019)Valldecabres, de~Benito, Casal, and
  Pablos}]{valldecabres2019design}
Valldecabres, R.; de~Benito, A.; Casal, C.; and Pablos, C. 2019.
\newblock DESIGN AND VALIDITY OF A BADMINTON OBSERVATION TOOL (BOT).
\newblock \emph{Revista internacional de medicina y ciencias de la actividad
  fisica y del deporte}, 19(74).

\bibitem[{Vaswani et~al.(2017)Vaswani, Shazeer, Parmar, Uszkoreit, Jones,
  Gomez, Kaiser, and Polosukhin}]{vaswani2017attention}
Vaswani, A.; Shazeer, N.; Parmar, N.; Uszkoreit, J.; Jones, L.; Gomez, A.~N.;
  Kaiser, {\L}.; and Polosukhin, I. 2017.
\newblock Attention is all you need.
\newblock \emph{Advances in neural information processing systems}, 30.

\bibitem[{Wang(2022)}]{wang2022modeling}
Wang, W.-Y. 2022.
\newblock Modeling Turn-Based Sequences for Player Tactic Applications in
  Badminton Matches.
\newblock In \emph{Proceedings of the 31st ACM International Conference on
  Information \& Knowledge Management}, 5128--5131.

\bibitem[{Wang et~al.(2022{\natexlab{a}})Wang, Chan, Peng, Yang, Wang, and
  Fan}]{wang2022stroke}
Wang, W.-Y.; Chan, T.-F.; Peng, W.-C.; Yang, H.-K.; Wang, C.-C.; and Fan, Y.-C.
  2022{\natexlab{a}}.
\newblock How Is the Stroke? Inferring Shot Influence in Badminton Matches via
  Long Short-Term Dependencies.
\newblock \emph{ACM Transactions on Intelligent Systems and Technology}, 14(1):
  1--22.

\bibitem[{Wang et~al.(2021)Wang, Chan, Yang, Wang, Fan, and
  Peng}]{wang2021exploring}
Wang, W.-Y.; Chan, T.-F.; Yang, H.-K.; Wang, C.-C.; Fan, Y.-C.; and Peng, W.-C.
  2021.
\newblock Exploring the long short-term dependencies to infer shot influence in
  badminton matches.
\newblock In \emph{2021 IEEE International Conference on Data Mining (ICDM)},
  1397--1402. IEEE.

\bibitem[{Wang et~al.(2022{\natexlab{b}})Wang, Shuai, Chang, and
  Peng}]{wang2022shuttlenet}
Wang, W.-Y.; Shuai, H.-H.; Chang, K.-S.; and Peng, W.-C. 2022{\natexlab{b}}.
\newblock Shuttlenet: Position-aware fusion of rally progress and player styles
  for stroke forecasting in badminton.
\newblock In \emph{Proceedings of the AAAI Conference on Artificial
  Intelligence}, volume~36, 4219--4227.

\bibitem[{Wang, Guo, and Zhao(2016)}]{wang2016badminton}
Wang, Z.; Guo, M.; and Zhao, C. 2016.
\newblock Badminton stroke recognition based on body sensor networks.
\newblock \emph{IEEE Transactions on Human-Machine Systems}, 46(5): 769--775.

\end{thebibliography}

\end{document}